\documentclass[letterpaper]{article} 
\usepackage[draft]{aaai2026}  
\usepackage{times}  
\usepackage{helvet}  
\usepackage{courier}  
\usepackage[hyphens]{url}  
\usepackage{graphicx} 
\usepackage{natbib}  
\usepackage{caption} 
\usepackage{algorithm}
\usepackage{algorithmic}
\usepackage{bm}

\usepackage{amssymb}
\usepackage{booktabs}

\usepackage[utf8]{inputenc}
\usepackage{amsmath}
\usepackage{tikz}
\usetikzlibrary{positioning}
\usetikzlibrary{calc}
\usepackage{caption}

\usepackage{cleveref}
\crefname{figure}{Fig.}{Figs.}
\crefname{table}{Tab.}{Tabs.}
\crefname{section}{Sec.}{Secs.}
\crefname{equation}{Eq.}{Eqs.}
\usepackage{newfloat}
\usepackage{listings}
\DeclareCaptionStyle{ruled}{labelfont=normalfont,labelsep=colon,strut=off} 
\floatstyle{ruled}
\newfloat{listing}{tb}{lst}{}
\floatname{listing}{Listing}
\usepackage{multirow}
\usepackage{xspace}

\definecolor{MyGray}{gray}{0.9}
\makeatletter
\newcommand{\thickhline}{%
    \noalign {\ifnum 0=`}\fi \hrule height 1pt
    \futurelet \reserved@a \@xhline
}

\newcommand{\subsmall}{\fontsize{8pt}{10pt}\selectfont}

\makeatletter
\DeclareRobustCommand\onedot{\futurelet\@let@token\@onedot}
\def\@onedot{\ifx\@let@token.\else.\null\fi\xspace}

\makeatother

\usepackage{mathtools}

\usepackage{caption} 
\title{Bayesian Neural Networks for One-to-Many Mapping in Image Enhancement}

\author{
    Guoxi Huang\textsuperscript{\rm 1},
    Qirui Yang\textsuperscript{\rm 2},
    Ruirui Lin\textsuperscript{\rm 1},
    Zipeng Qi\textsuperscript{\rm 3}, 
    David~Bull\textsuperscript{\rm 1},
    Nantheera Anantrasirichai\textsuperscript{\rm 1}
}
\affiliations{
    \textsuperscript{\rm 1} Visual Information Laboratory, 
 University of Bristol\\
     \textsuperscript{\rm 2} Tianjin University \\
     \textsuperscript{\rm 3} Baidu Inc \\

        \{guoxi.huang, r.lin, n.anantrasirichai, dave.bull\}@bristol.ac.uk, 
        yangqirui@tju.edu.cn, qizipeng1107@gmail.com

}

\begin{document}

\maketitle

\begin{abstract}
In image enhancement tasks, such as low-light and underwater image enhancement, a degraded image can correspond to multiple plausible target images due to dynamic photography conditions. This naturally results in a one-to-many mapping problem.
To address this, we propose a Bayesian Enhancement Model (BEM) that incorporates Bayesian Neural Networks (BNNs) to capture data uncertainty and produce diverse outputs. To enable fast inference, we introduce a BNN-DNN framework: a BNN is first employed to model the one-to-many mapping in a low-dimensional space, followed by a Deterministic Neural Network (DNN) that refines fine-grained image details.
Extensive experiments on multiple low-light and underwater image enhancement benchmarks demonstrate the effectiveness of our method.
\end{abstract}

\begin{links}
    \link{Code}{
    https://github.com/BinCVER/BEM}
\end{links}

\section{Introduction}

\begin{figure}[h]
    \begin{center}
    \includegraphics[width=.9\columnwidth]{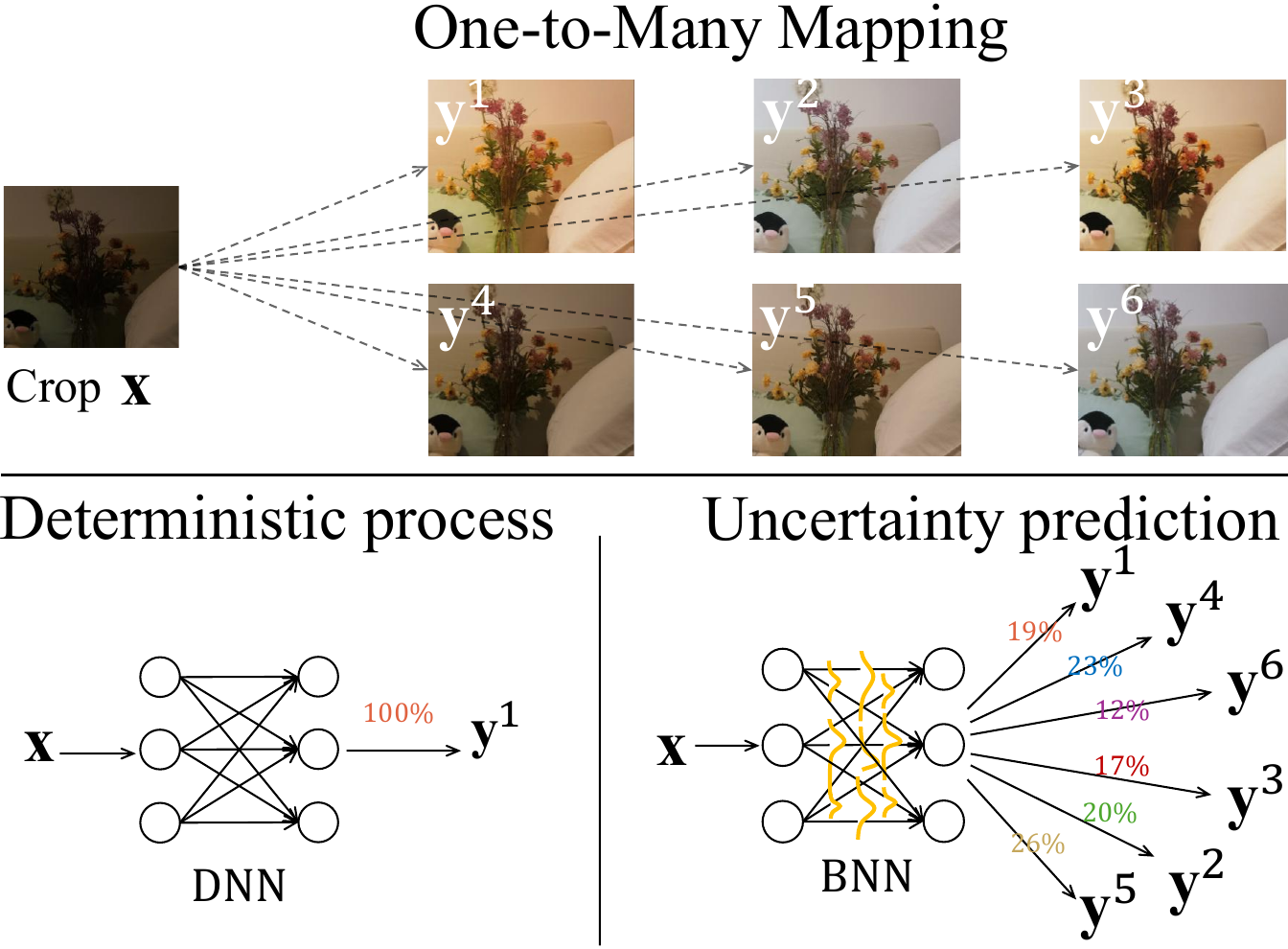}
    \caption{
     One-to-Many Mapping where an image crop \(\mathbf{x}\) associated with multiple targets \(\{\mathbf{y}^1, \ldots, \mathbf{y}^6\}\). A DNN (left) can only predict one of the targets. In contrast, a BNN (right) can produce many predictions according to a learned probability distribution.}
    \label{fig: one2many}
    \end{center}
\end{figure}

Image enhancement refers to the process of improving visual quality, primarily by adjusting illumination, as well as reducing noise, correcting colors, and refining structural details. The perceived quality of the enhanced image varies, as it is influenced by personal preferences and context-specific requirements.
In low-light image enhancement (LLIE) and underwater image enhancement (UIE) tasks, a significant challenge arises from the \emph{one-to-many mapping} problem, where a single degraded input image can correspond to multiple plausible target images.
As illustrated in~\cref{fig: one2many} (top), some reference images are unreliable due to poor visibility during image acquisition, which is often caused by challenging environments and limitations of imaging equipment.

Recent advances in deep learning have steered image enhancement towards data-driven approaches, with several models~\cite{peng2023u,cai2023retinexformer,li2021underwater} achieving state-of-the-art (SOTA) results by employing deterministic neural networks (DNNs) to learn one-to-one mappings between inputs and ground-truth images using paired datasets.
For LLIE and UIE tasks in particular, the ambiguity of target images makes DNNs poorly suited to capture the inherent variability in one-to-many image pairs, as illustrated in~\cref{fig: one2many} (left). 
In these enhancement tasks, ground-truth images are collected in real-world environments, which inevitably introduces low-quality targets into the training data—i.e., label noise. Such label noise is further amplified in extremely low-visibility data collection environments—particularly in challenging underwater or low-light scenes—where obtaining high-quality ground truth becomes nearly impossible.
As a result, learning a deterministic mapping from an input to a noisy ground truth can jeopardize enhancement quality.
However, directly discarding low-quality image pairs from training data leads to degraded performance in difficult scenes and harms the model’s generalization ability.

In this paper, we use a Bayesian Neural Network (BNN) to probabilistically model the one-to-many mappings between inputs and targets: We leverage Bayesian inference to sample network weights from a learned posterior distribution, where each sampled set of weights corresponds to a distinct plausible output. 
Through multiple sampling processes, the model maps a single input to a distribution of possible outputs, as illustrated in~\cref{fig: one2many} (right). Then, a ranking-based selection or Monte Carlo sampling is employed to obtain a reliable final output.

Although BNNs show promise in capturing uncertainty across various tasks\textcolor{black}{~\cite{kendall2016modelling,kendall2018multi}}, their potential for modeling the one-to-many mapping in image enhancement remains largely under-explored, despite clear benefits.
By incorporating Bayesian inference into the enhancement process, our approach captures uncertainty in dynamic, uncontrolled environments, providing a more flexible and robust solution than deterministic models.

Applying BNNs to high-resolution image tasks poses notable challenges: 1)~BNNs with high-dimensional weight spaces often suffer from underfitting~\cite{dusenberry2020efficient,tomczak2021collapsed}, hindering their ability to learn complex mappings effectively. To address this, we propose the \emph{Adaptive Prior}, which stabilizes training and accelerates convergence. 2)~Producing multiple high-resolution outputs with a BNN incurs substantial inference latency, making real-time processing impractical. To overcome this, we propose a two-stage BNN-DNN framework (Sec.~\ref{sec:A two-stage approach}) that captures one-to-many mappings in a compact low-dimensional space, significantly reducing computational cost while maintaining high-quality predictions.

We explore the feasibility of BNNs on the LLIE and UIE tasks where the \emph{one-to-many mapping} problem is particularly pronounced.
The main contributions of this paper are summarized as follows: 1) We identify the one-to-many mapping between inputs and targets as a key bottleneck in image enhancement models for LLIE and UIE, and propose a BNN-based method to address this challenge; 2) We design a two-stage BNN-DNN framework for efficient inference, enabling low-latency prediction by avoiding the explicit generation of multiple low-quality outputs. 3) We demonstrate that our method is backbone-agnostic and can benefit from future advances in backbone architectures.

\section{Background}

\paragraph{Bayesian Deep Learning.} BNNs quantify uncertainty by learning distributions over network weights, offering robust predictions~\cite{neal2012bayesian}. Variational Inference (VI) is a common method for approximating these distributions~\cite{blundell2015weight}.
\citet{gal2016dropout} simplified the implementation of BNNs by interpreting dropout as an approximate Bayesian inference method.
Another line of approaches, such as 
\citet{krishnan2020specifying}, explored the use of empirical Bayes to specify weight priors in BNNs to enhance the model's adaptability to diverse datasets. 
These BNN approaches have shown promise across a range of vision applications, including camera relocalization~\cite{kendall2016modelling}, semantic and instance segmentation~\cite{kendall2018multi}.
Despite these advances, BNNs remain underutilized in image enhancement tasks.

\paragraph{Visual Enhancement.} DNN-based methods~\cite{malyugina2025marine,huang2025bvi,zamir2022restormer} are widely used for image enhancement. Recently, probabilistic models have also been introduced to this domain. \citet{jiang2021enlightengan,islam2020fast} employed GANs for low-light and underwater image enhancement.
\citet{wang2022low} applied normalizing flow-based methods to reduce residual noise in LLIE predictions. However, its invertibility constraint limits model complexity. \citet{zhou2024glare} address the limitations of conventional normalizing flows by integrating a normal-light codebook with a latent normalizing flow, which more effectively aligns low-light and normal-light feature distributions.
Diffusion Models offer high fidelity and have been widely adopted for image enhancement tasks~\cite{hou2024global,tang2023underwater}, but they suffer from high inference latency due to their iterative denoising process.

\subsection{Preliminary}
In image enhancement, the output of a network can be interpreted as the conditional probability distribution of the target image, \(\mathbf{y} \in \mathcal{Y}\), given the degraded input image \(\mathbf{x} \in \mathcal{X}\), and the network’s weights \(\mathbf{w}\), i.e., \(P(\mathbf{y} \mid \mathbf{x}, \mathbf{w})\). Assuming the prediction errors follow a Gaussian distribution, the conditional probability density function of the target \(\mathbf{y}\) can be modeled as a multivariate Gaussian with mean given by the neural network output \(F(\mathbf{x}; \mathbf{w})\), i.e., \(P(\mathbf{y} \mid \mathbf{x}, \mathbf{w}) = \mathcal{N}(\mathbf{y} \mid F(\mathbf{x}; \mathbf{w}), \bm{\Sigma})\).

The network weights \(\mathbf{w}\) can be learned through maximum likelihood estimation (MLE). Given a dataset of image pairs \(\{\mathbf{x}^i, \mathbf{y}^i\}_{i=1}^N\), the MLE of $\mathbf{w}$, denoted as \(\mathbf{w}^{\mathrm{MLE}}\), is computed by maximizing the log-likelihood of the observed data:
\begin{equation}
\begin{aligned}
    \mathbf{w}^{\mathrm{MLE}} &= \underset{\mathbf{w}}{\operatorname{argmax}} \sum_{i=1}^N \log P(\mathbf{y}^i|\mathbf{x}^i, \mathbf{w}). \\ 
    \label{eq: mle}
\end{aligned}
\end{equation}

By optimizing such an objective function in~\cref{eq: mle}, the network $F_{\mathbf{w}}$ fits a one-to-one mapping, $F_\mathbf{w}: \mathcal{X} \rightarrow \mathcal{Y}$, implying that $\mathbf{y}^i \ne \mathbf{y}^j$ requires $\mathbf{x}^i \ne \mathbf{x}^j$. However, this assumption leads to mode collapse and poor expressiveness when modeling inherently one-to-many enhancement problems.

\begin{figure*}
\centerline{\includegraphics[width=.9\textwidth]{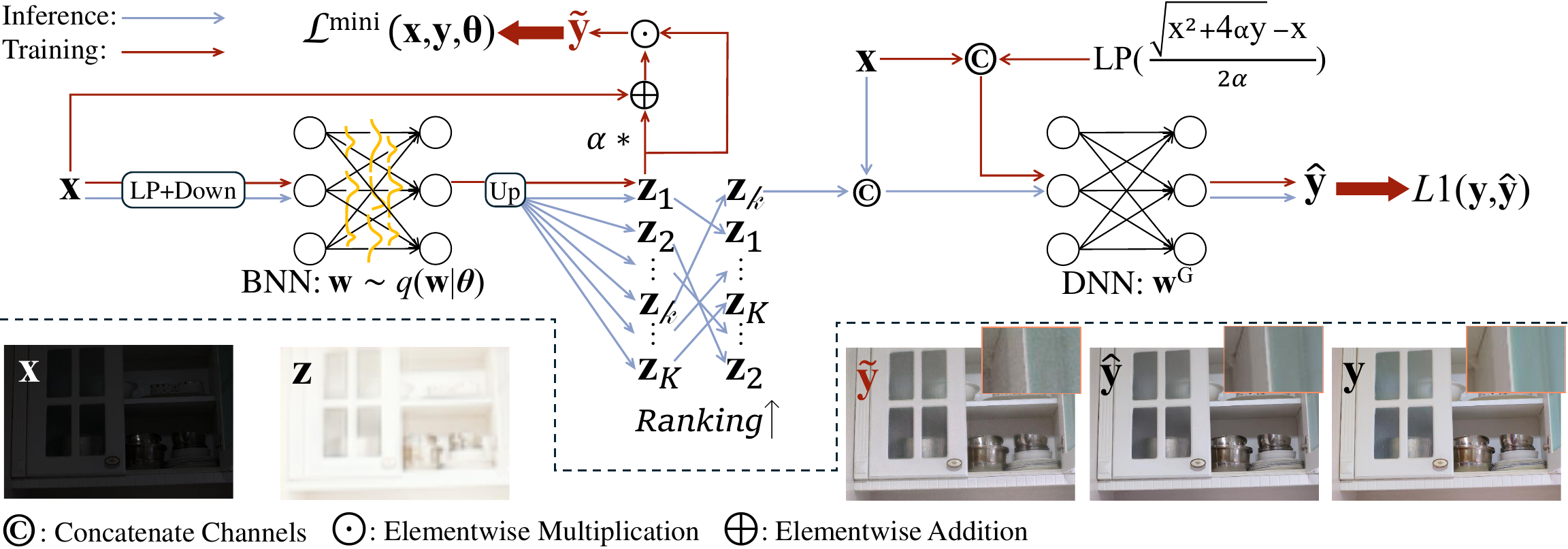}}
\caption{The two-stage pipeline. In Stage I, the BNN with weights $\mathbf{w} \sim q(\mathbf{w}|\bm{\theta})$ is trained by minimizing the minibatch loss $\mathcal{L}^\text{mini}$ in~\cref{eq:final_loss}. In Stage II, the DNN with weights $\mathbf{w}^\text{G}$ is trained by minimizing the L1 loss, $L1(\mathbf{y}, \hat{\mathbf{y}})$. The inference process is denoted by \textcolor{blue}{$\rightarrow$}, while the training process for each stage is indicated by \textcolor{red}{$\rightarrow$}.}
\label{fig:twostage}
\end{figure*}

\section{Method}
\label{sec:Modelling the one-to-many mapping}

\subsection{Variational Bayesian Inference for One-to-Many Modeling} 
\label{sec:Bayesian Enhancement Models}
To model the one-to-many mapping, we introduce uncertainty into the network weights $\mathbf{w}$ via Bayesian estimation, yielding a posterior distribution, $\mathbf{w} \sim P(\mathbf{w} | \mathbf{y}, \mathbf{x})$. During inference, weights are sampled from this distribution to generate diverse predictions. 
The posterior distribution over the weights is expressed as:
\begin{equation}
    \begin{aligned}
        P(\mathbf{w} | \mathbf{y}, \mathbf{x}) &=\frac{P(\mathbf{y} | \mathbf{x}, \mathbf{w} ) P(\mathbf{w})}{P(\mathbf{y} | \mathbf{x})},
    \end{aligned}
    \label{eq: posterior}
\end{equation}
where $P(\mathbf{y} \mid \mathbf{x}, \mathbf{w})$ is the likelihood of observing $\mathbf{y}$ given the input $\mathbf{x}$ and weights $\mathbf{w}$, $P(\mathbf{w})$ denotes the prior distribution of the weights, and $P(\mathbf{y} \mid \mathbf{x})$ is the marginal likelihood.

Unfortunately, for any neural networks, the posterior in~\cref{eq: posterior} cannot be calculated analytically. Instead, we can leverage variational inference (VI) to approximate $P(\mathbf{w} | \mathbf{y}, \mathbf{x})$ with a more tractable distribution $q(\mathbf{w} | \bm{\theta})$. Such that, we can draw samples of weights $\mathbf{w}$ from the distribution 
$q(\mathbf{w} | \bm{\theta})$. As suggested by~\cite{hinton1993keeping,blundell2015weight}, the variational approximation is fitted by minimizing their Kullback-Leibler (KL) divergence: 
\begin{equation}
    \begin{aligned}
    \bm{\theta}^\star &= \underset{\bm{\theta}}{\operatorname{argmin}}  
 ~\text{KL}\left[q(\mathbf{w} | \bm{\theta}) \Vert P(\mathbf{w} 
    | \mathbf{y}, \mathbf{x})\right] \\
    &= \underset{\bm{\theta}}{\operatorname{argmin}} \int q(\mathbf{w} | \bm{\theta}) \log \frac{q(\mathbf{w} | \bm{\theta})}{P(\mathbf{w})P(\mathbf{y} | \mathbf{x}, \mathbf{w} )} \, \mathrm{d} \mathbf{w}
    \\
    &= \underset{\bm{\theta}}{\operatorname{argmin}} -\mathbb{E}_{q(\mathbf{w}|\bm{\theta})}\left[\log P(\mathbf{y} | \mathbf{x}, \mathbf{w} )\right]  +
     \text{KL}\left[q(\mathbf{w} | \bm{\theta}) \Vert P(\mathbf{w})\right].
    \end{aligned}
    \label{eq: elbo}
\end{equation}
We define the resulting cost function from~\cref{eq: elbo} as:
\begin{equation}
    \begin{aligned}
    \mathcal{L}(\mathbf{x}, \mathbf{y}, \bm{\theta}) &=  \underbrace{-\mathbb{E}_{q(\mathbf{w}|\bm{\theta})}\left[\log P(\mathbf{y} | \mathbf{x}, \mathbf{w} )\right]}_{\text{data-dependent term}} 
     + \underbrace{\text{KL}\left[q(\mathbf{w} | \bm{\theta}) \Vert P(\mathbf{w})\right] }_{\text{prior matching term}},
    \end{aligned}
    \label{eq: loss_vfe}
\end{equation}
where the data-dependent term can be interpreted as a reconstruction error, such as \emph{L1} loss.

Let the variational posterior be a diagonal Gaussian $q(\mathbf{w}|\boldsymbol{\theta}) = \mathcal{N}(\boldsymbol{\mu}, \mathrm{diag}(\boldsymbol{\sigma}^2))$, parameterized by $\boldsymbol{\theta} = (\boldsymbol{\mu}, \boldsymbol{\sigma})$, which defines a BNN. During each forward pass of the BNN, the network weights $\mathbf{w}$ are sampled using the reparameterization trick~\cite{kingma2014auto}:
\begin{equation}
\mathbf{w} = \boldsymbol{\mu} + \boldsymbol{\sigma} \odot \boldsymbol{\epsilon}, \quad \text{where } \boldsymbol{\epsilon} \sim \mathcal{N}(0, \mathbf{I}).
\label{eq: reparam}
\end{equation}

\paragraph{Adaptive Prior.} 
To accelerate the convergence of Bayesian training, we adopt the idea of momentum update~\cite{he2020momentum} to establish an adaptive prior, which has been shown to achieve faster convergence than fixed or empirical priors. Specifically, the prior $P(\mathbf{w})$ at step $t$ is defined as:
\[
P(\mathbf{w}) = \mathcal{N}(\boldsymbol{\mu}_t^{\text{EMA}}, \mathrm{diag}((\boldsymbol{\sigma}_t^{\text{EMA}})^2)),
\]
where $\boldsymbol{\mu}_t^{\text{EMA}}$ and $\boldsymbol{\sigma}_t^{\text{EMA}}$ are updated via EMA from posterior parameters. This temporally adaptive prior smooths KL regularization across iterations, promoting consistent optimization. The resulting mini-batch loss follows:
\begin{equation}
\mathcal{L}^\mathrm{mini} = \frac{1}{M} \sum_i^M \mathbb{E}_{\mathbf{w} \sim q} \| F(\mathbf{x}^i; \mathbf{w}) - \mathbf{y}^i \|_2^2 
+ \text{KL}\left[q(\mathbf{w}) \| P(\mathbf{w}) \right],
\label{eq:final_loss}
\end{equation}

After optimizing \(\bm{\theta}^\star\) using~\cref{eq:final_loss}, the BNN generates multiple distinct predictions \(\{ \hat{\mathbf{y}}_1, \hat{\mathbf{y}}_2, \dots, \hat{\mathbf{y}}_K \}\) by sampling different weights \(\mathbf{w}\) from \(q(\mathbf{w} | \bm{\theta})\) during each forward pass.

\begin{figure*}[h]
\centerline{\includegraphics[width=1.\textwidth]{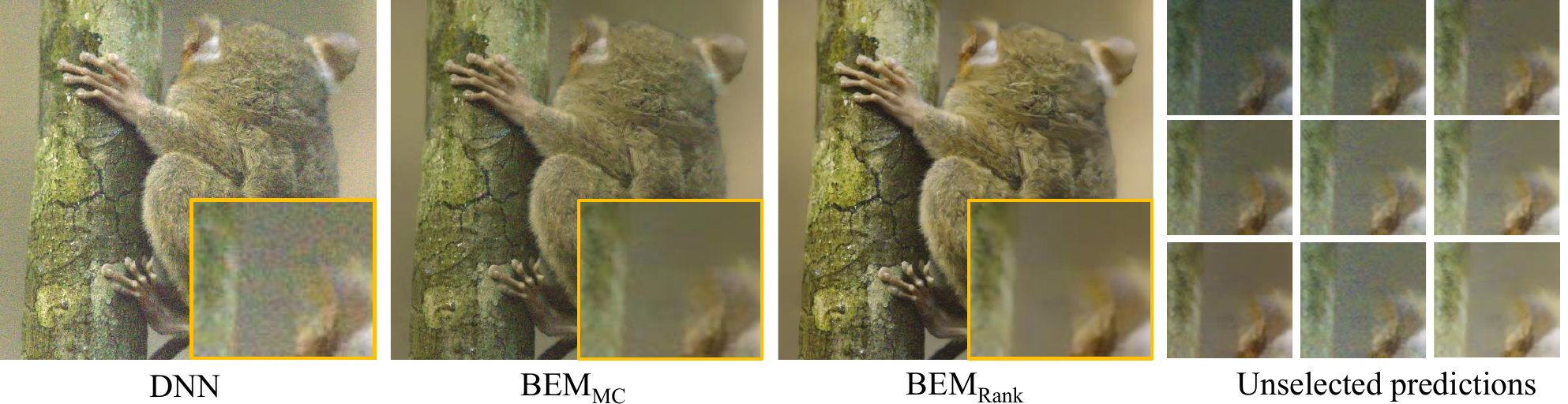}}
\caption{Visual comparisons of the DNN baseline, BEM$\text{MC}$, and BEM$\text{Rank}$ with CLIP-IQA. The rightmost patches highlight the diverse unselected predictions reflecting BEM’s one-to-many modeling capability.}
\label{fig:clip_cadidates}
\end{figure*} 

\subsection{BNN-DNN Framework}
\label{sec:A two-stage approach}

In a BNN, producing multiple high-resolution outputs can incur a high computational footprint. However, we are only interested in the highest-quality prediction among \(\{ \hat{\mathbf{y}}_1, \hat{\mathbf{y}}_2, \dots, \hat{\mathbf{y}}_K \}\).
To improve inference efficiency, we propose a two-stage architecture (see~\cref{fig:twostage}). The first stage uses a BNN to model one-to-many mappings in a low-dimensional latent space, capturing coarse structure and uncertainty. The second stage employs a DNN to reconstruct high-frequency details in the original image space.
The multiple coarse outputs \(\{\mathbf{z}_k\}_{k=1}^K\) from the first-stage BNN are used as proxies to identify the highest-quality prediction among \(\{\hat{\mathbf{y}}\}_{k=1}^K\), eliminating the need to produce all \(\{\hat{\mathbf{y}}\}_{k=1}^K\).

\textbf{In Stage I,} we employ low-pass filtering followed by downsampling to map the input's coarse information into a lower-dimensional space, \(\mathrm{Down}(\mathrm{LP}(\mathbf{x}), r)\), where 
$r$ denotes the scaling factor and $\mathrm{LP}$ represents a low-pass filter implemented via FFT. Subsequently, a BNN models the uncertainty in the low-dimensional coarse input information. The forward process of Stage I can be expressed as:
\begin{equation}
    \mathbf{z} = \mathrm{Up}(F(\mathrm{Down}(\mathrm{LP}(\mathbf{x}), r); \mathbf{w})), \quad \mathbf{w} \sim q(\mathbf{w} \mid \bm{\theta}), \label{eq:two_stageI} 
\end{equation}
where $\mathrm{Up}(\cdot)$ is the bilinear upsampling operation for dimensionality matching.
For a given input \(\mathbf{x}\), multiple proxies \(\{\mathbf{z}_k\}_{k=1}^K\) are generated via repeated forward passes in~\cref{eq:two_stageI}. The most reliable proxy $\mathbf{z}_k$ can then be automatically selected using a ranking mechanism (see~\cref{sec:Predictions Under Uncertainty}).
From~\cref{fig:twostage}, we can observe that \(\mathbf{z}\) approximates the enhanced illumination. The first-stage prediction \(\tilde{\mathbf{y}}\) in the original resolution is computed as: 
\begin{equation}
   \tilde{\mathbf{y}} = (\mathbf{x} + \alpha \mathbf{z}) \odot  \mathbf{z},
\label{eq:predy_stageI} 
\end{equation}
where \(\alpha\) is a small scalar and $\odot$ is element-wise multiplication. Compared to simpler formulations, such as \(\mathbf{x} + \mathbf{z}\) or \(\mathbf{x} \odot \mathbf{z}\), \cref{eq:predy_stageI} reduces the risk of blurring fine textures or amplifying noise in \(\mathbf{x}\). We note that \(\tilde{\mathbf{y}}\) plays a key role in the ranking-based inference in~\cref{sec:Predictions Under Uncertainty}.

\textbf{In Stage II,} we employ a DNN $G$ to enhance the fine-grained details in the input. The forward process can be expressed as:
\begin{equation}
    \hat{\mathbf{y}} = G([\mathbf{x}, \mathbf{z}]; \mathbf{w}^\mathrm{G}), \label{eq:two_stageII}
\end{equation}
where \(\mathbf{w}^\mathrm{G}\) represents the weights of the second-stage model, $[\cdot,\cdot]$ denotes the concatenation operation along the channel dimension.    
When training the second-stage DNN, we replace the predicted coarse information \(\mathbf{z}\) with its ground truth, \(\mathrm{LP}\left(\frac{\sqrt{\mathbf{x}^2 + 4\alpha\mathbf{y}} - \mathbf{x}}{2\alpha}\right)\), which is the explicit solution of~\cref{eq:predy_stageI} when \(\tilde{\mathbf{y}}\) is replaced with \(\mathbf{y}\) and \(\mathbf{z}\) is treated as the unknown variable.
This strategy is critical, as it prevents mode collapse—where diverse predictions from the first-stage BNN are undesirably regressed into a single output by the second-stage DNN.

\paragraph{The Backbone.}
For both the first- and second-stage models, we adopt the same backbone network but use different input and output layers.  
In the first stage, we construct a BNN by converting all layers in the backbone to their Bayesian counterparts via~\cref{eq: reparam}. 
The backbone follows an encoder-decoder UNet design. For the basic blocks, we consider both Transformers~\cite{vaswani2017attention} and Mamba~\cite{gu2023mamba}, demonstrating the broad applicability of our methods across the two primary backbone architectures.
We provide more details in the Supplementary Materials.

\subsection{Inference under Uncertainty}
\label{sec:Predictions Under Uncertainty}

As described in Algorithm~\ref{algo:infer}, our method enables two inference modes: a ranking-based selection and Monte Carlo (MC) sampling.
In the first stage, the BNN generates \(K\) latent candidates \(\{\mathbf{z}_k\}_{k=1}^K\), which are computed in parallel.
For ranking-based inference, an image quality assessment metric, \(\mathsf{IQA}(\cdot)\), is applied to score the intermediate predictions \(\{\tilde{\mathbf{y}}_k\}_{k=1}^K\) from~\cref{eq:predy_stageI}.
The latent code \(\mathbf{z}^*\) corresponding to the highest-ranked \(\tilde{\mathbf{y}}_k\) is selected and passed to the second stage for refinement, yielding the final output.

\begin{algorithm}
\small
\caption{Inference}
\label{algo:infer}
\begin{algorithmic}
\REQUIRE Input $\mathbf{x}$, BNN $F$, DNN $G$
\FOR{$k = 1$ to $K$}
    \STATE $\mathbf{w}_k \gets \bm{\mu} + \bm{\sigma} \circ \bm{\epsilon}_k$, \hspace{0.8cm} where $\bm{\epsilon}_k \sim \mathcal{N}(\mathbf{0}, \mathbf{I})$
    \STATE $\mathbf{z}_k \gets F(\mathrm{Down}(\mathrm{LP}(\mathbf{x}), r); \mathbf{w}_k)$ \hfill \textit{ \large  \textcolor{gray}{ Stage I}}
\ENDFOR
\IF{Mode = \textit{Monte Carlo}}
    \STATE $\mathbf{z}^* \gets \frac{\mathbf{z}_1 + \mathbf{z}_2 + \cdots + \mathbf{z}_K}{K}$
\ELSE
    \STATE $\mathbf{z}^* \gets \underset{{\mathbf{z}_k \in \{\mathbf{z}_1, \mathbf{z}_2, \dots, \mathbf{z}_K\}}}{\operatorname{argmax}} \mathsf{IQA}((\mathbf{x} + \alpha \mathbf{z}_k) \circ \mathbf{z}_k)$
\ENDIF
\STATE $\hat{\mathbf{y}} \gets G([\mathbf{x}, \mathrm{Up}(\mathbf{z}^*)]; \mathbf{w}^\mathrm{G})$ \hfill \textit{ \large  \textcolor{gray}{  Stage II}}
\ENSURE $\hat{\mathbf{y}}$
\end{algorithmic}
\end{algorithm} 

\begin{figure}[h]
\centerline{\includegraphics[width=1.\columnwidth]{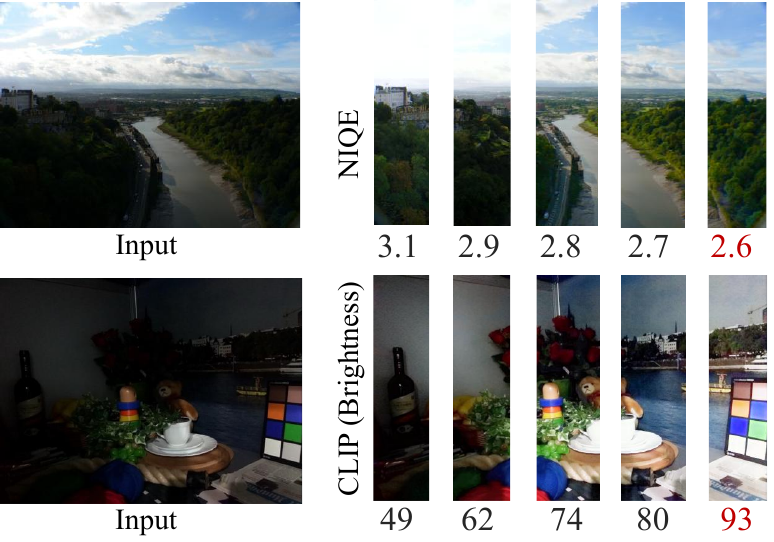}}
\caption{One-to-many mapping from input to outputs. The predictions are sorted by CLIP-IQA and NIQE. 
}
\label{fig:visual_predict}
\end{figure}

For MC inference, we aggregate the \(K\) latent samples \(\{\mathbf{z}_1, \dots, \mathbf{z}_K\}\) by computing their mean \(\mathbf{z}^*\), which is subsequently passed to the second-stage DNN to generate the final prediction \(\hat{\mathbf{y}}\).
This process approximates the posterior expectation over the BNN’s stochastic predictions. Since the BNN implicitly captures label noise through weight uncertainty, averaging multiple samples serves to marginalize out the randomness induced by noisy supervision, resulting in more stable and noise-suppressed outputs.

We denote the variant that uses ranking-based inference as BEM$_\text{Rank}$ and the one that applies Monte Carlo sampling as BEM$_\text{MC}$. The ranking mode can be instantiated with various no-reference IQA metrics, including CLIP-IQA, NIQE, UIQM, and UCIQE.

As illustrated in~\cref{fig:visual_predict}, CLIP-based ranking tends to favor brighter, high-contrast outputs that align with semantic and perceptual cues learned from natural images, while NIQE emphasizes statistical naturalness. 
\cref{fig:clip_cadidates} further shows that both BEM$_\text{Rank}$ and BEM$_\text{MC}$ generate visually enhanced results with minimal noise, whereas the deterministic DNN baseline exhibits notable residual artifacts. 
Due to its averaging nature, BEM$_\text{MC}$ produces conservative outputs with smoothed details, whereas BEM$_\text{Rank}$ often yields higher-contrast results with enhanced perceptual sharpness.
Although automatic ranking improves robustness and efficiency, users may also manually select their preferred enhancement from multiple candidates when speed is not a primary concern.

\begin{figure}
\flushleft
\includegraphics[width=1.\columnwidth]{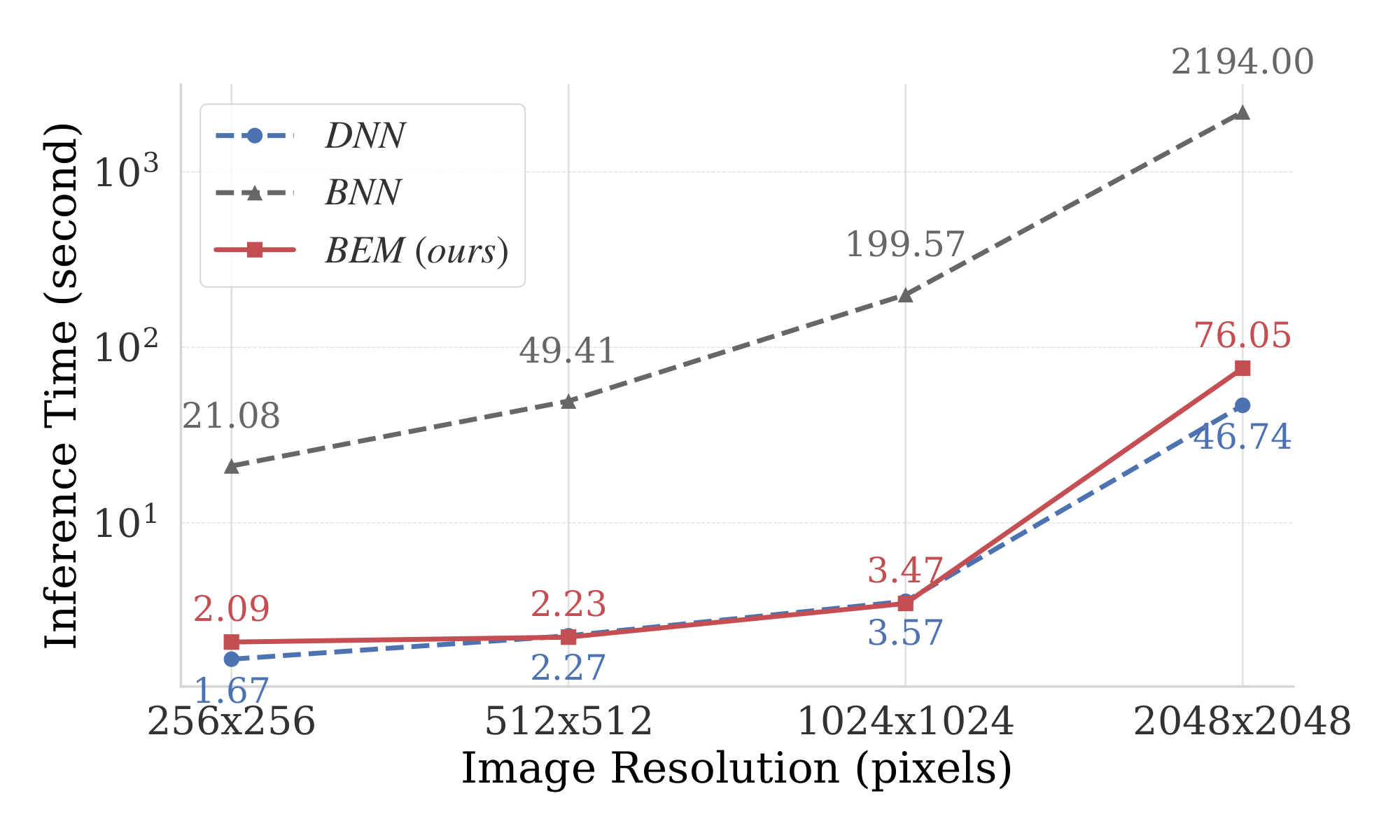}
\caption{Inference speed on an Nvidia RTX 4090.}
\label{fig:runtime}
\end{figure}

\paragraph{Inference Speed.}
Algorithm~\ref{algo:infer} avoids redundant sampling, resulting in a substantial reduction in inference latency. 
\cref{fig:runtime} compares the inference time of a conventional BNN, a standard DNN, and the proposed two-stage BEM. BEM achieves runtime comparable to the DNN and delivers a 22$\times$ speedup over the BNN when processing $512 \times 512$ images.

\begin{figure}[h]
\centerline{\includegraphics[width=1.\columnwidth]{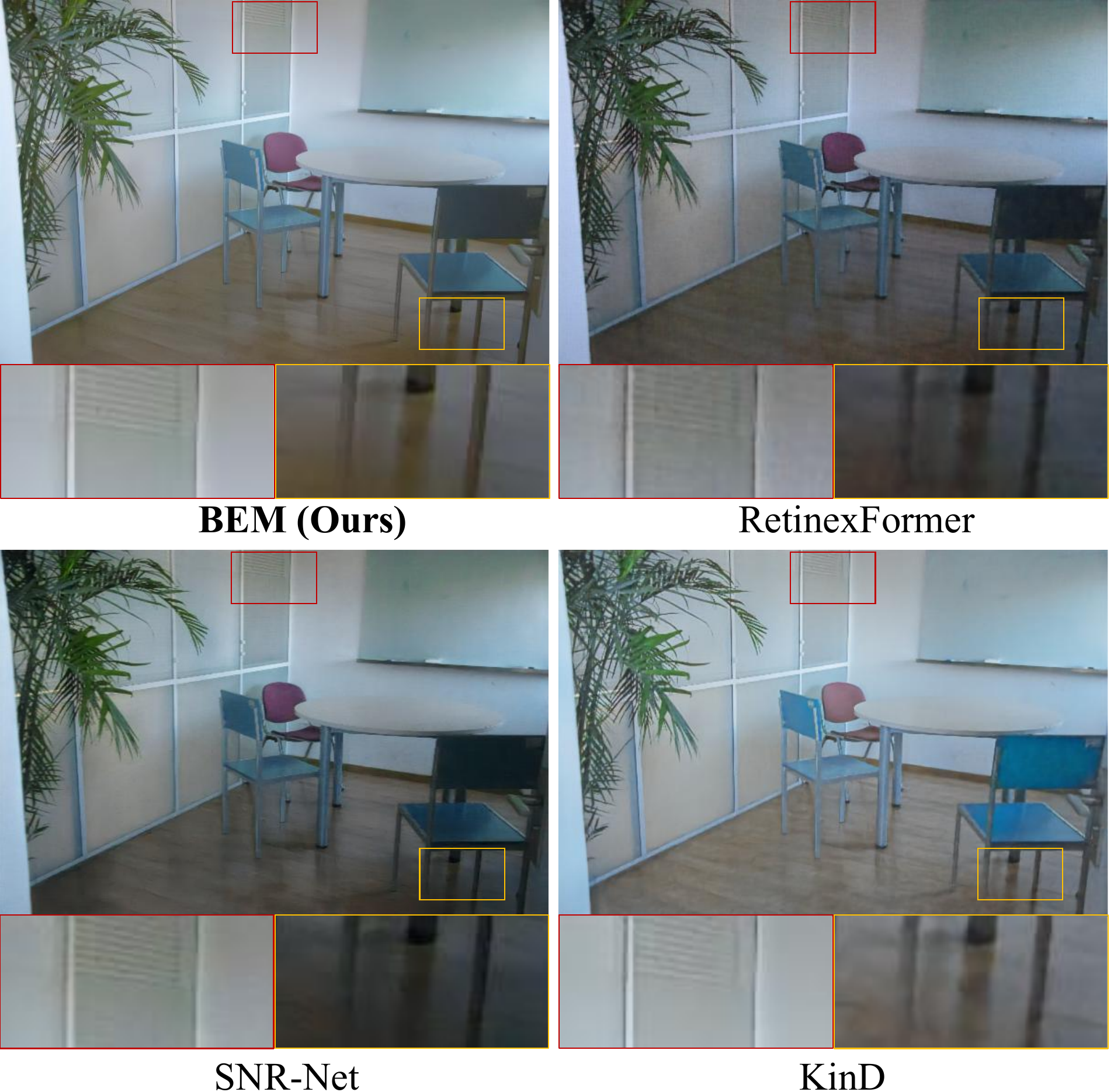}}
\caption{Visual comparisons on the LOL dataset.}

\label{fig:visual_lol_zoomin}
\end{figure}

\section{Experiments}
\label{sec:Experiments}

\paragraph{Datasets.} 
For LLIE, we evaluate our method on the paired LOL-v1~\cite{wei2018deep} and LOL-v2~\cite{yang2021sparse} datasets, as well as the unpaired LIME~\cite{guo2016lime}, NPE~\cite{wang2013naturalness}, MEF~\cite{ma2015perceptual}, DICM~\cite{lee2013contrast}, and VV~\cite{vonikakis2018evaluation} datasets. For UIE, we evaluate our method on the paired UIEB-R90~\cite{li2019underwater} dataset, along with unpaired datasets including C60 and U45~\cite{li2019fusion}.

\begin{table*}[h]
    \centering
    \subsmall
    \setlength{\tabcolsep}{8pt}
    \begin{tabular}{@{}ll|ccccccccc@{}}
    \toprule
        \multirow[c]{2}{*}{ \small Method}  & \multirow[c]{2}{*}{\small Type} & \multicolumn{3}{c}{LOL-v1}  & \multicolumn{3}{c}{LOL-v2-real}  & \multicolumn{3}{c}{LOL-v2-syn}  \\
        \addlinespace[-2pt]
        \cmidrule(lr){3-5} \cmidrule(lr){6-8} \cmidrule(lr){9-11}
        \addlinespace[-2pt]
       & & PSNR $\uparrow $ & SSIM $\uparrow $& LPIPS $\downarrow$ & PSNR $\uparrow $ & SSIM $\uparrow $& LPIPS $\downarrow$ & PSNR $\uparrow $ & SSIM $\uparrow $& LPIPS $\downarrow$  \\
        \midrule
        \textcolor{gray}{\textit{Without GT Mean}} \\
        KinD  & DNN & 19.66 &  0.820 & 0.156 & 18.06 & 0.825 & {0.151} &  17.41&  0.806 & 0.255 \\
        Restormer & DNN  & 22.43&  0.823 & 0.147 & 18.60  & 0.789 & 0.232 & 21.41 & 0.830 & 0.144  \\
        SNR-Net  & DNN& 24.61 &  0.842 & 0.151 & 21.48  & {0.849} &  0.157 & 24.14 &  0.928 & 0.056 \\ 
        RetinexFormer &DNN & {25.16} & {0.845}  &  {0.131} & {22.80} & 0.840 &  0.171  & {25.67} & {0.930} &  {0.059} \\
        RetinexMamba & DNN & 24.03 & 0.827 & 0.146 & 22.45 & 0.844 & 0.174 & {25.89} & {0.935} & 0.054 \\

  Mamba $\mid$ $\text{BEM}_\text{MC}$ (ours)& BNN & 23.07 & 0.851 & 0.089 & 20.13 & 0.858 & 0.107 & 24.57 & 0.938 & 0.043 \\
        
        \midrule

        \textcolor{gray}{\textit{With GT Mean}} \\
        {LLFlow} & Flow & {25.13} & {0.872} & {0.117} & {26.20} &{0.888} & {0.137} & {24.81} & {0.919} & {0.067}  \\
        EnlightenGAN & GAN & 17.48 & 0.652 & 0.322  & 18.64 & 0.677 & 0.309 & 16.57  & 0.734 & - \\
        {{GlobalDiff}}   & DM &  {\underline{27.84}} & {{0.877}} & {{0.091}} & {{28.82}} & {{0.895}}  & {{0.095}} & {{28.67}}  & {{0.944}} & {{0.047}}  \\
        {{GLARE}}  & Flow & {27.35} & {\textbf{0.883}} & {0.083} & {28.98}  & {0.905} & {0.097} & {29.84}  & {{0.958}}  & {-}\\
        
        \textcolor{gray}{Transformer} $\mid$ \textcolor{gray}{{$\text{BEM}_\text{Rank}$ (ours)}}  & BNN &  \textcolor{gray}{{{28.24}}} & \textcolor{gray}{{{0.881}}} & \textcolor{gray}{{{0.077}}} &  \textcolor{gray}{{{32.54}}} & \textcolor{gray}{{{0.917}}} & \textcolor{gray}{{{0.072}}} & \textcolor{gray}{{32.36}} & \textcolor{gray}{{0.962}} & \textcolor{gray}{{0.030}} \\
        
        \textcolor{black}{Transformer} $\mid$ 
 {$\text{BEM}_\text{MC}$ (ours)} & BNN & {{27.22}} & {0.879} & 
        {{0.075}} &  {\underline{30.86}} &  {{0.905}} &  {\underline{0.069}}  & {{30.21}} & {{0.944}} & {{0.035}} \\ 
        
        \textcolor{gray}{Mamba} $\mid$ \textcolor{gray}{{$\text{BEM}_\text{Rank}$ (ours)}} & BNN  & \textcolor{gray}{{{28.80}}} & \textcolor{gray}{{{0.884}}} & \textcolor{gray}{{{0.069}}} &  \textcolor{gray}{{{32.66}}} & \textcolor{gray}{{{0.915}}} & \textcolor{gray}{{{0.060}}} & \textcolor{gray}{{32.95}} & \textcolor{gray}{{0.964}} & \textcolor{gray}{{0.026}} \\
        
        \textcolor{black}{Mamba} $\mid$ 
 {$\text{BEM}_\text{MC}$ (ours)} & BNN & {\textbf{28.30}} & {0.881} & 
        {\underline{0.072}} &  {\textbf{31.41}} &  {\textbf{0.912}} &  {\textbf{0.064}}  & {\underline{30.58}} & {\underline{0.958}} & {\underline{0.033}} \\ 

    \bottomrule%
    \end{tabular}
    \caption{Full-reference evaluation on LOL-v1 and v2. The best results are in \textbf{bold}, while the second-best are \underline{underlined}. Results in \textcolor{gray}{gray} indicate the upper bound performance of BEM and are not directly comparable to the other results.
    }
    \label{tab:result_LLIE}
\end{table*}

\noindent \textbf{Experimental Settings.} All models are trained using the Adam optimizer, with an initial learning rate of $2 \times 10^{-4}$, decayed to $10^{-6}$ following a cosine annealing schedule. The first- and second-stage models are trained for 300K and 150K iterations, respectively, on $128 \times 128$ inputs with a batch size of \(M = 8\).
Unless stated otherwise, the downscale factor $r$ in \cref{eq:two_stageI} is set to \(\frac{1}{16}\), \(K\) to 25, $\alpha$ in~\cref{eq:predy_stageI} to 0.025, and the adopted backbone architecture is Mamba.

\subsection{Evaluation}
\label{sec: full_ref_eval}

For paired test sets with reliable ground-truth images, we evaluate enhancement quality using full-reference metrics, including PSNR, SSIM, and LPIPS. 
In unpaired or real-world scenarios where reference images are unavailable, we report no-reference scores using NIQE, UIQM, and UCIQE.

\begin{table}[!t]
    \centering
    \subsmall
    \setlength{\tabcolsep}{0.5pt}
    \begin{tabular}{@{}ll|cc|cccccc@{}}
    \toprule
        \multirow{2}{*}{\small Method} & \multirow{2}{*}{\small Type} & \multicolumn{2}{c}{UIEB-R90} & \multicolumn{2}{c}{C60}   &  \multicolumn{2}{c}{U45} \\
        \addlinespace[-2pt]
        \cmidrule(lr){3-4} \cmidrule(lr){5-6} \cmidrule(lr){7-8}
        \addlinespace[-2pt]
        & & PSNR $\uparrow$ & SSIM $\uparrow$ & UIQM $\uparrow$ & UCIQE $\uparrow$ & UIQM $\uparrow$ & UCIQE $\uparrow$ \\
        \midrule
        Ucolor     & DNN  & 20.13 & 0.877 &  2.482 & 0.553  & 3.148 &  0.586 \\
        PUIE-MP    & VAE  & 21.05 & 0.854 &  2.524 &  0.561  & 3.169 & 0.569 \\
        Restormer  & DNN  & 23.82 & 0.903 & 2.688  & \underline{0.572} & 3.097 & 0.600 \\
        FUnIEGAN   & GAN  & 19.12 &  0.832 & \underline{2.867} & 0.556 &  2.495  & 0.545 \\
        PUGAN      & GAN  & 22.65 &  0.902 & 2.652 & 0.566 & - & - \\ 
        U-Shape    & DNN  & 20.39 & 0.803  & 2.730 & 0.560 & 3.151 &  0.592 \\
        Semi-UIR   & DNN  & 22.79 & \underline{0.909} & 2.667 & \textbf{0.574} & {3.185} & 0.606 \\
        WFI2-Net   & DNN  & \
        \underline{23.86} & 0.873 & - & - & 3.181 & \underline{0.619} \\
        \midrule
        \textcolor{black}{BEM$_\text{MC}$}   & BNN & \textbf{23.92} & \textbf{0.913} &  {2.694} & {0.537} & \underline{3.188} & {0.591} \\
        BEM$_{\text{Rank}}$   & BNN & \textcolor{gray}{25.62} & \textcolor{gray}{0.940} & \textbf{2.931} &  0.567 & \textbf{3.406} &  \textbf{0.620} \\
    \bottomrule
    \end{tabular}
    \caption{Full-reference evaluation (left) on R90, and no-reference evaluations (right) on C60 and U45.}
    \label{tab:result_UIE}
\end{table}

\begin{table}
    \centering
    \subsmall
    \setlength{\tabcolsep}{8pt}
    \begin{tabular}{@{}ll|ccccc@{}}
    \toprule
    {Method} & Type  & DICM  & LIME  & MEF & NPE & VV  \\
    \midrule
    KinD      & DNN   & 5.15 & 5.03 & 5.47 &  4.98 & 4.30 \\
    ZeroDCE   & DNN   & 4.58 & 5.82 & 4.93 & 4.53 &  4.81 \\
    RUAS      & DNN   & 5.21  &  4.26 & 3.83 & 5.53 & 4.29 \\
    LLFlow    & Flow  & 4.06 & 4.59 & 4.70 & 4.67 & 4.04 \\
    PairLIE   & DNN   & 4.03 & 4.58 & 4.06 & 4.18 & 3.57 \\
    RFR       & DNN   & {3.75} & \underline{3.81} &3.92  &4.13 & - \\
    GLARE     & Flow  & \underline{3.61} & 4.52  & 3.66  & 4.19  & - \\ 
    CIDNet    & DNN   & 3.79  & 4.13 &  3.56  & \underline{3.74}  & 3.21 \\
    \midrule
    $\text{BEM}_\text{MC}$  & BNN & \textcolor{black}{3.77} & \textcolor{black}{3.94} & \underline{3.22} & \textcolor{black}{3.85} & \underline{2.95} \\
    $\text{BEM}_\text{Rank}$  & BNN &  \textbf{3.55} & \textbf{3.56} &  \textbf{3.14} & \textbf{3.72} & \textbf{2.91} \\
    \bottomrule
    \end{tabular}
    \caption{No-reference NIQE$\downarrow$ evaluation, compared to previous methods~\cite{yan2025hvi,guo2020zero,liu2021retinex,fu2023learning,fu2023you}}
    \label{tab:result_no_ref_LLIE}
\end{table}

We compare against various types of probabilistic models and leading DNN-based methods, including normalizing flows~\cite{wang2022low,zhou2024glare}, GANs~\cite{jiang2021enlightengan,cong2023pugan,islam2020fast}, diffusion models~\cite{hou2024global}, variational autoencoders (VAEs)~\cite{fu2022uncertainty}, as well as strong deterministic baselines~\cite{zhang2019kindling,zamir2022restormer,xu2022snr,cai2023retinexformer,bai2024retinexmamba}.

\paragraph{Full-reference.}
We conduct full-reference comparisons on LLIE (LOL-v1/v2) and UIE (UIEB-R90), as reported in \cref{tab:result_LLIE} and \cref{tab:result_UIE} (middle). 
Our BEM, equipped with either a Transformer or Mamba backbone, achieves superior performance across all metrics and datasets, highlighting its robustness and generalization across diverse conditions. 
In contrast to prior methods that struggle to balance perceptual quality (e.g., LPIPS) and pixel-level fidelity (e.g., SSIM), BEM achieves higher SSIM and lower LPIPS simultaneously.  
For full-reference evaluation, we report BEM$_\text{Rank}$ as an upper-bound performance of our method, since its first-stage ranking selects the candidate that is closest to the reference image under the chosen metric.

\paragraph{No-reference.}
We further evaluate LLIE performance on five unpaired datasets using no-reference metrics. As shown in~\cref{tab:result_no_ref_LLIE}, both $\text{BEM}_\text{MC}$ and $\text{BEM}_\text{Rank}$ achieve lower (i.e., better) NIQE scores than other methods. \cref{fig:visual_predict} shows that outputs with lower NIQE scores generally exhibit more natural illumination and avoid overexposure, indicating that $\text{BEM}_\text{Rank}$ with NIQE-based ranking can reliably identify high-quality predictions in LLIE. \cref{tab:result_UIE} (right) further demonstrates that BEM achieves the best or competitive performance on the two unpaired UIE benchmarks.

\begin{figure*}
\centerline{\includegraphics[width=1.\textwidth]{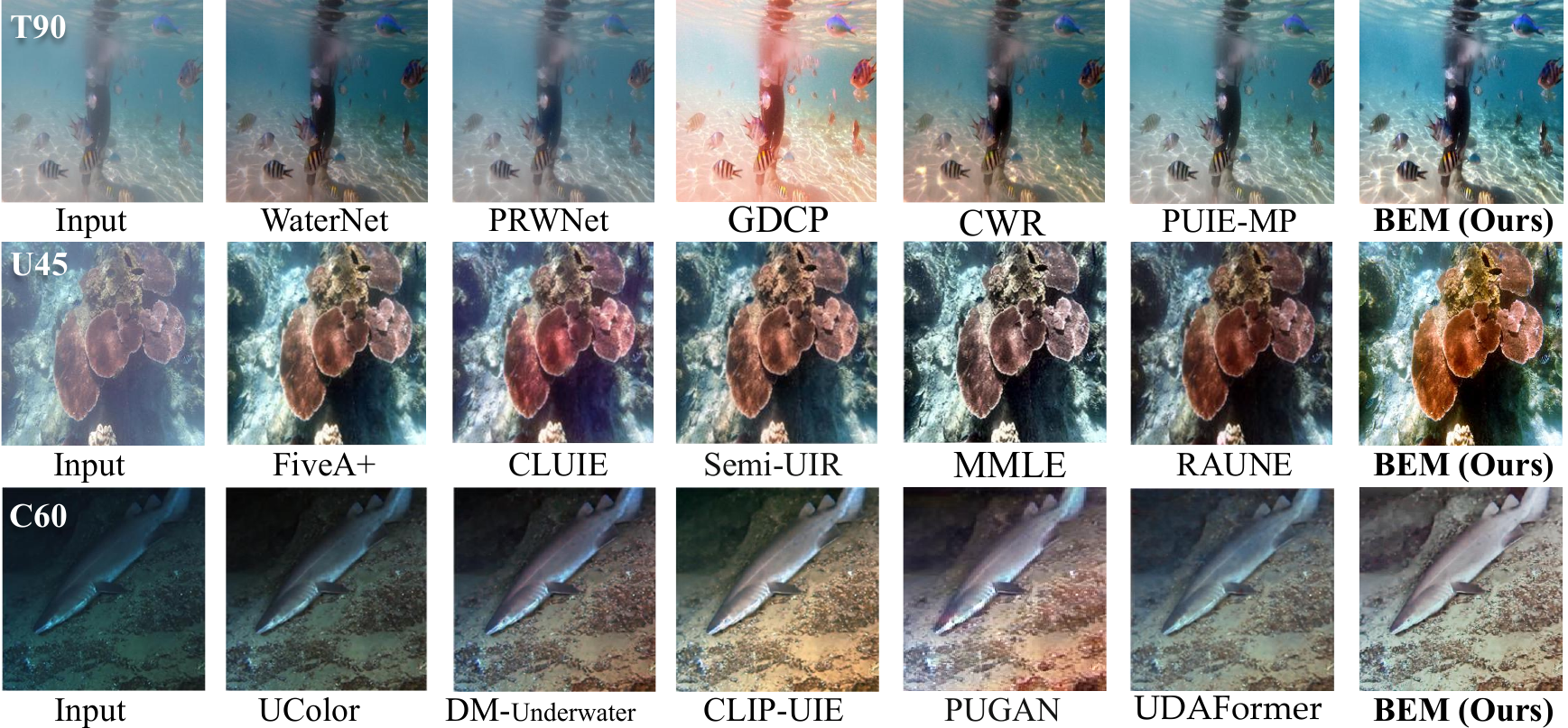}}
\caption{Visual comparisons
on the R90, C60 and U45 datasets. 
Best viewed when zoomed in. 
} 
\label{fig:visual_UIE}
\end{figure*}

\paragraph{Qualitative Results.}
As shown in~\cref{fig:visual_lol_zoomin}, our BEM better preserves fine details and structural textures, as evidenced by the zoomed-in regions, compared to existing methods.
\cref{fig:visual_UIE} further compares our method with representative UIE approaches, including~\citet{tang2023underwater,fu2022uncertainty,islam2020fast,cong2023pugan,fu2022uncertainty,li2019underwater,jiang2023five,li2021underwater,tang2023underwater,huo2021efficient,zhang2022underwater,huang2023contrastive,liu2024underwater,li2022beyond,peng2018generalization,shen2023udaformer,han2021single}.
Visual comparisons suggest that BEM restores color and illumination more faithfully, producing more natural-looking outputs, particularly in challenging scenes such as C60.

\begin{figure}[h]
\centerline{\includegraphics[width=1.\columnwidth]{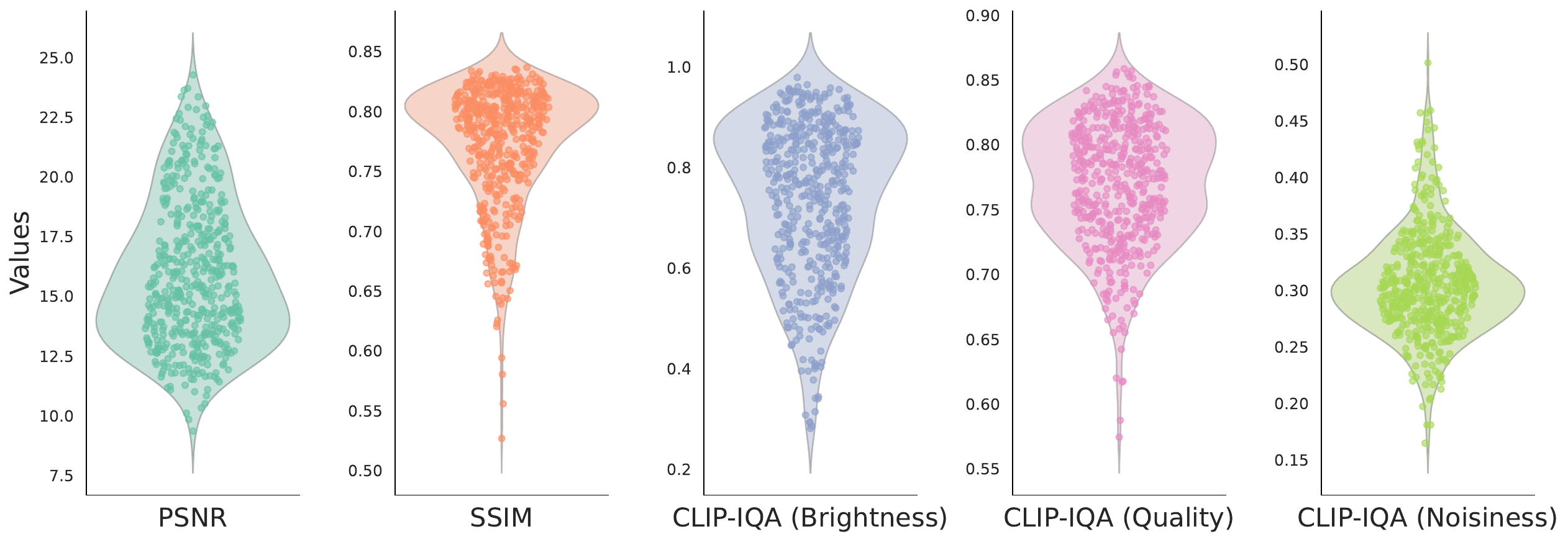}}
\caption{Score distributions of 500 predictions from BEM across PSNR, SSIM, and three CLIP-IQA metrics (Brightness, Quality, Noisiness).}
\label{fig:violinplot}
\end{figure}
\begin{figure}
\centerline{\includegraphics[width=1.\columnwidth]{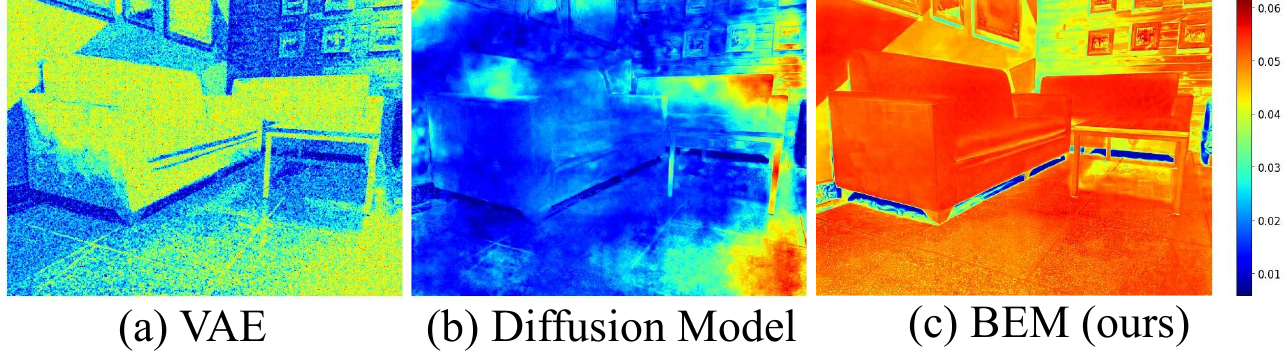}}
\caption{
Visualization of pixel-wise output variability from 500 samples for: (a) VAE, (b) Diffusion, and (c) our BEM. Brighter regions indicate higher uncertainty.
}
\label{fig:pixel_std}
\end{figure}

\begin{figure}
\vspace{-5pt}
\centerline{\includegraphics[width=1.\columnwidth]{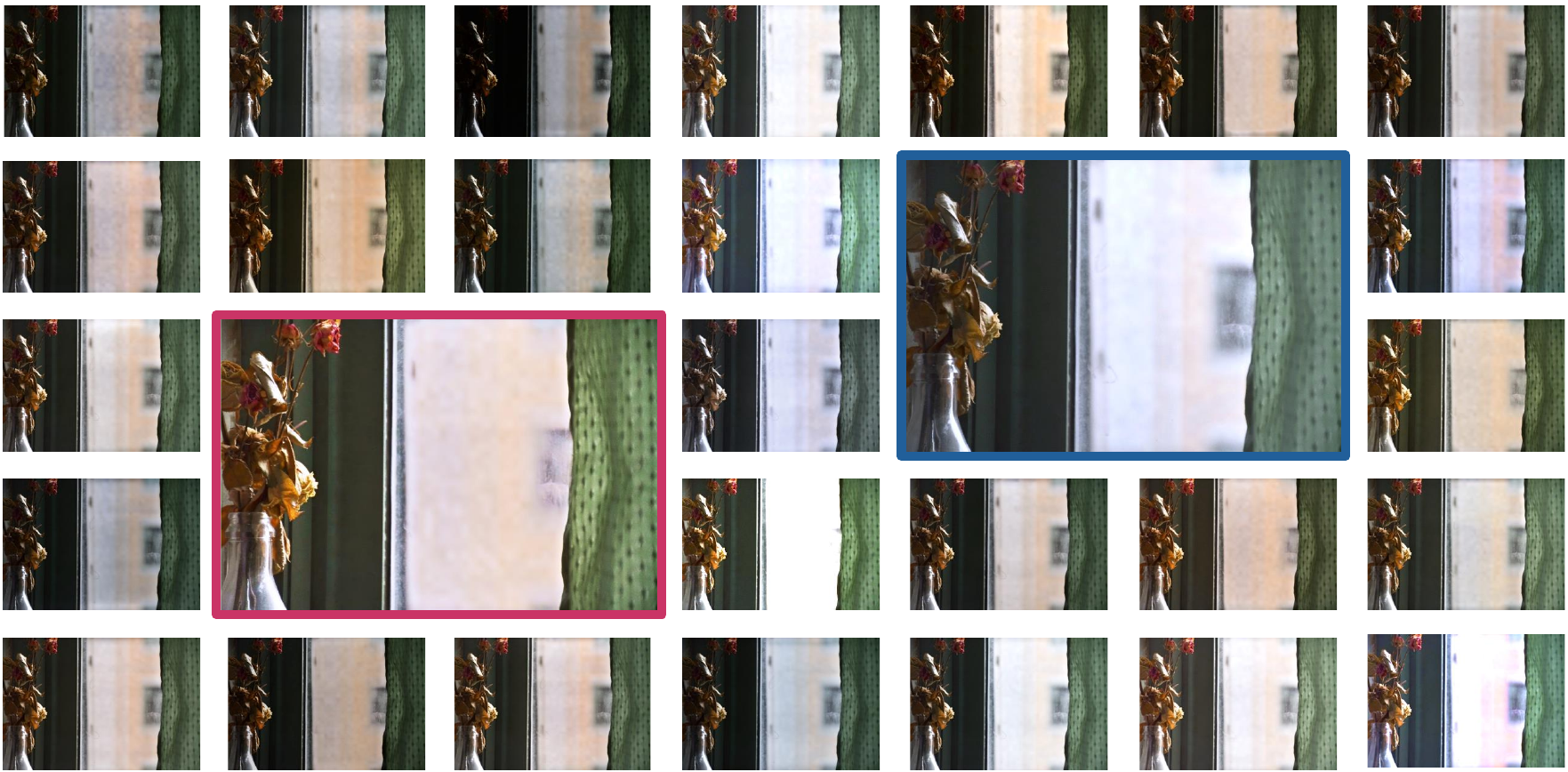}}
\caption{Visualization of BEM predictions. The input image is from LSRW~\citep{hai2023r2rnet}.}
\label{fig:clip_cadidates2}
\end{figure}

\paragraph{Statistical Analysis of Uncertainty}
\cref{fig:violinplot} shows the score distributions of predictions from BEM. The spread in these distributions reflects the predictive variability inherent in one-to-many modeling. Notably, a portion of the samples receives low scores under multiple metrics, which indicates the presence of noisy or low-quality labels in the training set. This observation supports the inclusion of uncertainty modeling to better accommodate imperfect supervision.

\paragraph{BEM vs. Diffusion Model and VAE.}
\label{abl: Uncertainty Maps}
To assess the one-to-many modeling capacity of different probabilistic approaches, we visualize the pixel-wise output variability of the VAE, the diffusion model, and our BEM under identical inputs. As shown in~\cref{fig:pixel_std}, BEM produces substantially higher output variability while still preserving localized structural consistency, particularly along object boundaries where the predicted uncertainty remains low. In comparison, both the conditional VAE and the diffusion model display markedly lower variability, revealing a trade-off between generative diversity and structural fidelity. These observations indicate that BEM can maintain sample diversity without sacrificing geometric alignment, which is essential for modeling one-to-many mappings in ill-posed enhancement tasks. \cref{fig:clip_cadidates2} further illustrates the diverse enhanced outputs produced by BEM, all of which exhibit consistent structural appearance.

\section{Conclusion}

We presented the Bayesian Enhancement model (BEM) to address the one-to-many challenge in image enhancement, which is identified as a key limitation in previous data-driven models.  An \textit{Adaptive Prior} is introduced to support stable and efficient Bayesian training, while the BNN–DNN design enables fast inference.
Experiments across multiple benchmarks show clear improvements over existing methods, highlighting the benefits of Bayesian modeling for ambiguous enhancement tasks.

\section*{Acknowledgments}
This work was supported by the UKRI MyWorld Strength
 in Places Program (SIPF00006/1) and the EPSRC ECR
 International Collaboration Grants (EP/Y002490/1). We acknowledge Humble Bee Films for providing visual content.
\bibliography{main}

\end{document}